\documentclass[sigconf]{acmart}

\usepackage{epsfig}
\usepackage{graphicx}
\usepackage{amsmath}
\usepackage{amssymb}
\usepackage{multirow}

\newcommand{\tabincell}[2]{\begin{tabular}{@{}#1@{}}#2\end{tabular}} 

\usepackage{mathrsfs}
\usepackage{subfigure}

\usepackage{paralist} 

\newenvironment{packed_itemize}{
	\begin{itemize}
		\setlength{\itemsep}{0pt}
		\setlength{\parskip}{0pt}
		\setlength{\parsep}{0pt}
	}{\end{itemize}}

\AtBeginDocument{%
  \providecommand\BibTeX{{%
    \normalfont B\kern-0.5em{\scshape i\kern-0.25em b}\kern-0.8em\TeX}}}

\copyrightyear{2020}
\acmYear{2020}
\setcopyright{acmcopyright}\acmConference[MM '20]{Proceedings of the 28th ACM
	International Conference on Multimedia}{October 12--16, 2020}{Seattle, WA, USA}
\acmBooktitle{Proceedings of the 28th ACM International Conference on Multimedia
	(MM '20), October 12--16, 2020, Seattle, WA, USA}
\acmPrice{15.00}
\acmDOI{10.1145/3394171.3413849}
\acmISBN{978-1-4503-7988-5/20/10}

\begin{document}
\fancyhead{}
\title{Compositional Few-Shot Recognition with Primitive Discovery and Enhancing}

\author{Yixiong Zou$^{1,5}$, Shanghang Zhang$^2$, Ke Chen$^3$, Yonghong Tian$^{1*}$, Yaowei Wang$^4$, Jos\'e M. F. Moura$^5$}
\thanks{$^*$ indicates corresponding author.}
\affiliation{%
\institution{Peking University$^1$, University of California, Berkeley$^2$, South China University of Technology$^3$}
\institution{PengCheng Laboratory$^4$, Carnegie Mellon University$^5$}
}
\email{{zoilsen, yhtian}@pku.edu.cn, shz@eecs.berkeley.edu, chenk@scut.edu.cn, wangyw@pcl.ac.cn, moura@andrew.cmu.edu}

\renewcommand{\shortauthors}{Yixiong Zou, et al.}

\begin{abstract}

Few-shot learning (FSL) aims at recognizing novel classes given only few training samples, which still remains a great challenge for deep learning. However, humans can easily recognize novel classes with only few samples.
A key component of such ability is the compositional recognition that human can perform, which has been well studied in cognitive science but is not well explored in FSL.
Inspired by such capability of humans, to imitate humans' ability of learning visual primitives and composing primitives to recognize novel classes, we propose an approach to FSL to learn a feature representation composed of important primitives, which is jointly trained with two parts, i.e. primitive discovery and primitive enhancing.
In primitive discovery, we focus on learning 
primitives related to object parts by self-supervision from the order of image splits, avoiding extra laborious annotations and alleviating the effect of semantic gaps.
In primitive enhancing, 
inspired by current studies on the interpretability of deep networks,
we provide our composition view for the FSL baseline model. 
To modify this model for effective composition,
inspired by both mathematical deduction and biological studies (the Hebbian Learning rule and the Winner-Take-All mechanism), we propose a soft composition mechanism by enlarging the activation of important primitives while reducing that of others, so as to enhance the influence of important primitives and better utilize these primitives to compose novel classes. 
Extensive experiments on public benchmarks are conducted on both the few-shot image classification and video recognition tasks. 
Our method achieves the state-of-the-art performance on all these datasets and shows better interpretability.

\end{abstract}



\maketitle

\begin{figure}[t]
	\centering
	\includegraphics[width=0.9\columnwidth, height=0.45\columnwidth]{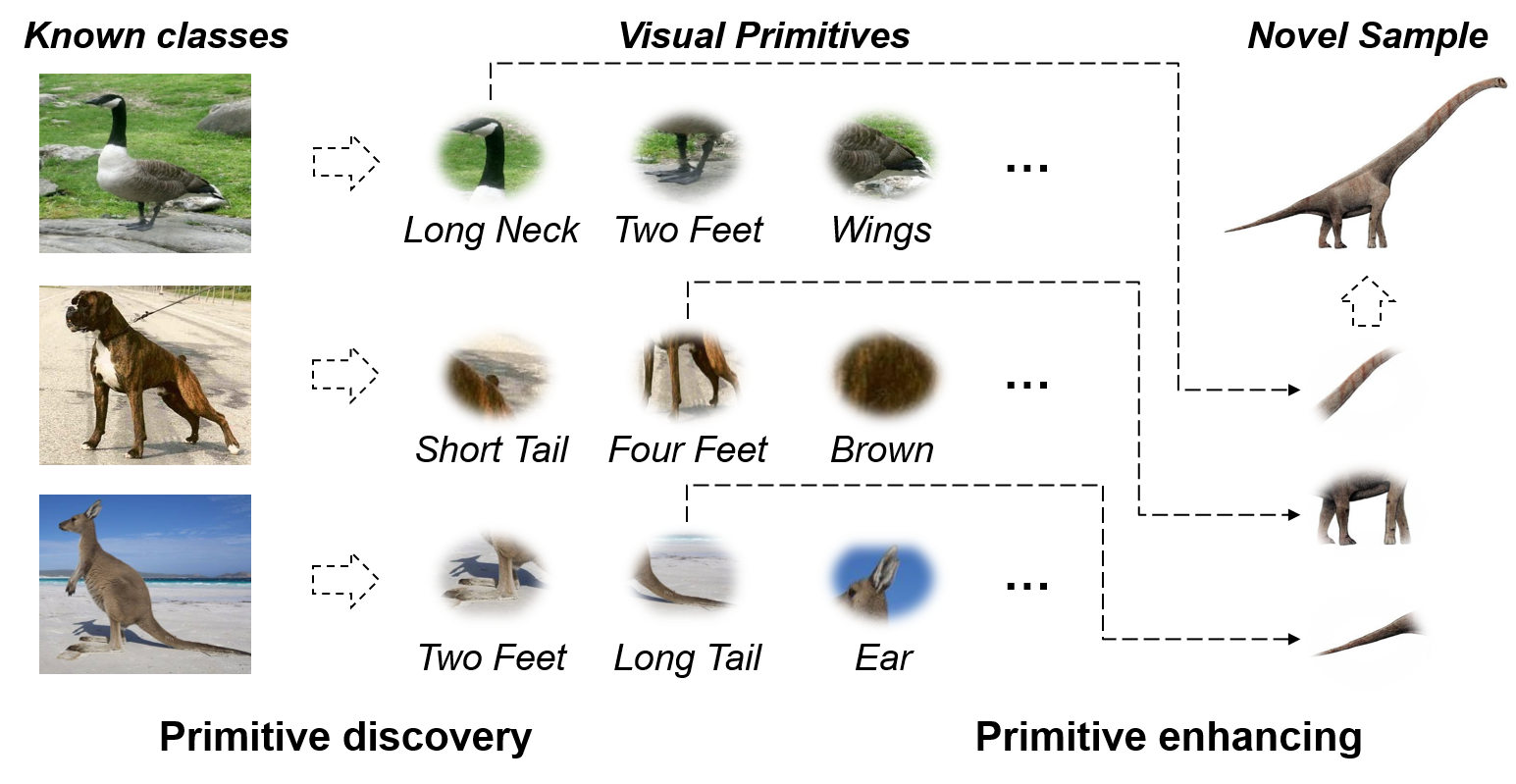}\vspace{-0.35cm}
	\caption{ Human can decompose known classes into primitives and use the composition of learned primitives to recognize novel samples. To imitate this ability, we propose an approach to few-shot recognition to learn a feature representation composed of important primitives, which is jointly trained with primitive discovery and primitive enhancing.}
	\label{fig: motivation}\vspace{-0.55cm}
\end{figure}

\vspace{-0.4cm}
\section{Introduction}


Recently, deep learning has achieved superior performance in various tasks with sufficient labeled data.
However, in practice, labels in visual recognition are expensive to obtain via manual annotation, and new classes of objects may arise dynamically in nature. Therefore, it is extremely difficult to annotate sufficient samples for these new classes. 
To address these limitations, few-shot learning (FSL) has been researched actively in recent years and recognized as a feasible solution \cite{ravi2016optimization}, which categorizes objects from novel classes using only few training samples, with prior knowledge transferred from non-overlapping known classes that have sufficient data.
However, there is still a big gap between machines and humans in the recognition ability. Humans can recognize novel classes with only few samples. As studied in cognitive science, a key component of such ability is the compositional recognition~\cite{biederman1987recognition}, which means humans can first learn primitives~\cite{hoffman1984parts} from known classes and then compose novel concept with the learned primitives~\cite{fodor1975language}, as shown in Fig.\ref{fig: motivation}.
In practice, primitives are viewed as object parts, or more broadly, components capturing the compositional structure of the examples~\cite{tokmakov2019learning}.
Although known classes and novel classes are non-overlapping, they can share some primitives in common.
The concept of compositional recognition has been applied in some domains such as VQA~\cite{andreas2016learning} and human-object interaction~\cite{kato2018compositional}. However, such concept has not been well explored in FSL. 
In this paper, to imitate humans' ability of learning primitives and utilizing primitives to compose novel classes, we propose an approach to FSL to compose novel-class samples with important primitives from known classes, which is jointly trained with two parts, i.e. primitive discovery and primitive enhancing.


In primitive discovery, 
we propose to use the self-supervision from object split orders to facilitate the discovery of part-related primitives and reduce the effect of semantic gaps.
Specifically,
as visual primitives can refer to object parts~\cite{tokmakov2019learning}, the training procedure should encourage the model to recognize object parts. However, as current methods mainly rely on the supervision from image-level class labels, it may be harder to achieve this goal than that under the explicit supervision of object parts. 
Also, annotating all possible object parts is prohibitively expensive.
Moreover, in the compositional recognition, transferring primitives learned on known classes to novel classes assumes they are semantically related. However, various classes with different semantic gaps may exist in novel classes.
Therefore, to assist the discovery of part-related primitives without laborious annotations and reduce the effect of semantic gaps, we propose to use the self-supervision from object split orders to assist the learning procedure.
Specifically, we split the input image horizontally and vertically, perm the splits, and ask the model to recognize which perms are applied to them. 
As the splitting operation tends to break the entire object into parts, the model is encouraged to recognize splits by recognizing parts, thus discovering primitives and latently encoding them in the network.
Moreover, as objects with relatively large semantic gaps may share similar structures (e.g., the upper and lower parts of dogs and cars can be easily distinguished, although they are not highly semantically related), self-supervision from object structures~\cite{noroozi2016unsupervised, kim2019self} may help to alleviate the influence of semantic gaps.
As the supervision from class labels still dominants the training, primitives other than object parts will not be sacrificed.

In primitive enhancing, we propose the soft composition mechanism with an Enlarging-Reducing loss (ER loss) to compose novel-class samples with the discovered primitives and enhance the influence of important primitives.
Specifically, our method is based on the widely adopted baseline model~\cite{li2019large, DBLP:journals/corr/abs-1904-04232, hariharan2017low, qiao2017few, rusu2019meta, tokmakov2019learning}, which first conducts known-class classification with deep networks, then utilizes the trained network to extract feature for novel-class samples, and finally performs the Nearest Neighbor classification.
Current works~\cite{bau2017network, fong2018net2vec, zhou2016learning} on the deep networks' interpretability show that channels in the penultimate layer of deep networks can correspond to some certain patterns (e.g., texture, object parts) in the input sample.
Inspired by these works, we propose the soft composition mechanism based on the \textit{compositional view} which regards each channel of the penultimate layer as a primitive. 
During known-class training, the classification probability of the input can be treated as the normalized weighted sum of the activation on primitives, where the weights of primitives are represented in the fully connected (FC) layer's parameters. 
During novel-class testing, as all channels are already contained in the extracted novel-class feature, the cosine-similarity-based Nearest Neighbor classification can be viewed as comparing two primitive sets and outputting an overall similarity score.
Since the novel-class composition of the baseline model suffers from the problem that all primitives are averagely weighted, which harms the composition by the influence from those trivial primitives, a straightforward solution may be the hard composition mechanism, which explicitly select all the important primitives and neglect those trivial ones. However, such mechanism may require to modify the network structure to allow a dynamic number of primitives, which is complicated and difficult to implement.
Therefore, to both achieve effective composition and simplify the hard composition mechanism, we design the soft composition mechanism with the ER loss, which enhances the influence of important primitives. 
With higher influence from these important primitives, the unknown-class features can be better composed. 
The proposed ER loss is inspired both mathematically and biologically.
Mathematically, important primitives in known-class recognition should have larger influence on novel-class recognition. Therefore, we first define the influence of primitives in the novel-class classification. Then by deduction, we find that the influence of those important primitives can be enhanced by enlarging the activation of them while reducing that of others during the known-class training, leading to the proposed ER loss.
Biologically, as each primitive is weighted by the parameter (neuron) in the FC layer connected to it, this primitive-neuron pair can be viewed as two connected cells. As empirically highly-activated FC neurons are always connected to highly-activated primitives, according to the biologically widely observed Hebbian Learning rule~\cite{hebb1949organization}, we enlarge the activation of the connected primitives, which can be implemented as the enlarging term of the ER loss. Moreover, as primitives can be viewed as competing with each other to get a higher importance during known-class training, according to the Winner-Take-All (WTA) mechanism~\cite{douglas2004neuronal} in human cortex, we reduce the activation of the losers' activation, which can be implemented as the reducing term of the ER loss.
Although the ER loss is simple, it is biologically interpretable and empirically effective, and requires no extra parameters or modifications to the network structure.
Moreover, combining with the compositional view, we can then explain which primitives from known classes compose the given novel sample by the feature map visualization, showing better interpretability for the deep-learning-based FSL.
After the above training on known classes, the trained network will be used to extract features and perform Nearest Neighbor classification on novel classes.

Compared with current works, our model has better interpretability due to both the compositional recognition (as validated in section~\ref{sec: Experiments: visualization}) and the biological interpretation of the ER loss (as stated in section~\ref{sec:bio_ER}).
The most relevant work with ours is \cite{tokmakov2019learning}, which pushes the visual feature close to the sum of human annotated attributes. Compared with it, our method differs in (1) we don't need human annotated attributes, (2) instead of feature vectors, we treat channels within a single feature layer as the visual primitives, and the composition is achieved in the channel level instead of the feature level (summing feature vectors), (3) our model is easier to be jointly trained in an end-to-end manner, while \cite{tokmakov2019learning} points out that for easier converging, it must adopt a two-stage training strategy.

In all, our contributions can be summarized as follows.
\vspace{-0.3cm}
\begin{packed_itemize}
	\item Inspired by humans' compositional recognition, we propose an approach to FSL to learn a feature representation composed of important primitives, which is jointly trained with primitive discovery and primitive enhancing.
	\item To facilitate the discovery of part-related primitives without laborious annotations and reduce the effect of semantic gaps, in primitive discovery, we propose to use the self-supervision from object split orders to assist the primitive learning.
	\item To compose the novel-class feature, in primitive enhancing, we provide our compositional view for the FSL baseline model. To modify this model for effective composition, inspired both mathematically and biologically (the Hebbian learning rule and the WTA mechanism), we propose a soft composition mechanism by enlarging the activation of important primitives while reducing that of others (ER loss).
	\item Extensive experiments on three popular benchmarks demonstrate better interpretability and superior performance of the proposed method compared to the state-of-the-arts on both few-shot image and video recognition tasks.
\end{packed_itemize}

\vspace{-0.3cm}
\section{Related Work}

\textbf{Few-shot learning} (FSL) methods can typically be grouped into meta learning based methods, metric learning based methods, and data augmentation based methods. 
The meta-learning based methods develop a meta-learner model that can quickly adapt to a new task given a few training examples \cite{andrychowicz2016learning,munkhdalai2017meta,finn2017model,grant2018recasting,lee2018gradient,zhang2018metagan,ravi2016optimization}.
Typical works include learning model initializations~\cite{finn2017model}, learning the stochastic gradient decent optimizer~\cite{ravi2016optimization} and learning the weight-update mechanism with an external memory~\cite{munkhdalai2017meta}. 
The embedding and metric learning based methods address the FSL problem by learning the feature representations that preserve the class neighborhood structure and comparing samples \cite{koch2015siamese,mensink2012metric,snell2017prototypical,Vinyals2016Matching,yang2018learning,garcia2017few}. 
The data augmentation based methods solve the FSL problem by augmenting the training data using prior knowledge, such as learning a data generator to hallucinate novel-class data \cite{hariharan2017low,wang2018low}. 
However, they mainly treat each class as a whole, while we decompose them into primitives and operate on the primitive level to select and enhance the most effective ones, leading to a better interpretability.

\noindent\textbf{Compositional recognition} is the recognition by primitives, and has been well studied in the cognitive science~\cite{biederman1987recognition, hoffman1984parts, fodor1975language}. This concept has been applied to some domains: \cite{misra2017red} points that the model will benefit from compositional learning of visual tasks to have better generalization to novel tasks. \cite{kato2018compositional} proposes to decompose human-object interaction into action and object.
\cite{purushwalkam2019task} decomposes complex attributes into simple ones, and learns their compositions for zero-shot learning. 
Very recently \cite{tokmakov2019learning} proposes to push the visual feature close to the combination of attribute embedded features, decomposing the visual feature into manual labeled attributes.
However, this concept is still far from being well explored in FSL. All these methods largely rely on human annotated attributes/database (e.g. CUB~\cite{wah2011caltech} attributes used in \cite{tokmakov2019learning}), which is expensive. Compared with them, our method is able to learn without laborious annotations, and can be easily trained in an end-to-end manner.

\noindent\textbf{Interpretability of deep networks} has been studied in many aspects~\cite{bau2017network, fong2018net2vec, zhou2016learning}, showing that channels in the penultimate layer of deep networks correspond to some certain patterns patterns in the input images. These patterns may include color, texture, object, etc~\cite{bau2017network}. 
In this work, inspired by these studies, we view each channel as a primitive, and developed a soft composition mechanism to compose novel-class features by learned primitives.

\noindent\textbf{Self-supervised learning} aims at learning from the supervision of the object structure, alleviating the need of supervision from manual labels, and has been researched in the field of unsupervised/semi-supervised learning~\cite{noroozi2016unsupervised, kim2019self}. Recently, this mechanism is also applied in FSL by methods such as predicting the rotation~\cite{mangla2019charting,gidaris2019boosting} and predicting the relative position~\cite{gidaris2019boosting}. In this work, inspired by these previous works, we propose to use the self-supervised split loss to learn part-related primitives and alleviate the influence of the semantic gap between known and novel classes.

\vspace{-0.1cm}
\section{Methodology}

\begin{figure}[t]
	\centering\includegraphics[width=0.95\columnwidth, height=0.55\columnwidth]{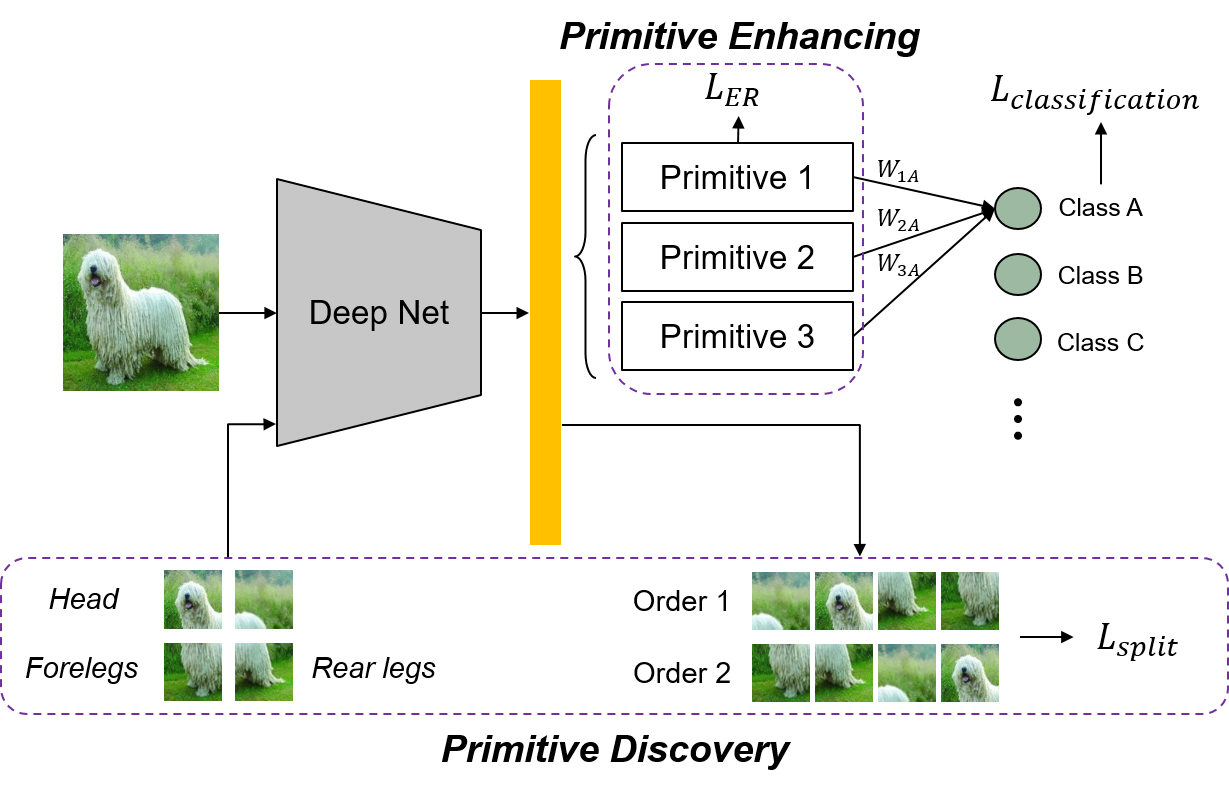}\vspace{-0.35cm}
	\caption{Framework. A self-supervised loss based on the split order prediction is applied to discover part-related primitives and alleviate the effect of semantic gaps. To better utilize learned primitives to compose novel classes, a soft composition mechanism is proposed, which is achieved by the math- and bio-inspired ER loss to enlarge the activation of important primitives while reducing that of others.}
	\label{fig: framework}\vspace{-0.35cm}
\end{figure}

In this section, the proposed method aims to imitate humans' compositional recognition, and is jointly trained with primitive discovery and primitive enhancing. A self-supervised split loss is applied to discover the primitives related to object parts without laborious annotations and alleviate the effect of semantic gaps. To compose novel classes with learned primitives, we provide our compositional view for the FSL baseline model. To modify this model for effective composition, a soft composition mechanism is proposed, which is achieved by the math- and bio-inspired Enlarging-Reducing (ER) loss to enlarge the activation of important primitives while reducing that of others. The framework is shown in Fig.\ref{fig: framework}. The network is trained on known classes, and then the Nearest Neighbor classification is performed on novel classes.

\vspace{-0.1cm}
\subsection{FSL baseline model}
\label{sec: preliminaries}

Following the setting of current works~\cite{Vinyals2016Matching,ravi2016optimization}, we are provided with a large-scale known set $\mathcal{D}_{known}$ with known classes $\mathcal{C}_{known}$ and a novel/unknown\footnote{\textit{Novel class} is equivalent to \textit{Unknown class}.} set $\mathcal{D}_{unknown}$ with unknown classes $\mathcal{C}_{unknown}$. Note that $\mathcal{C}_{known} \bigcap \mathcal{C}_{unknown} = \emptyset$. Few-shot learning aims at recognizing query samples from unknown classes given only few (1 to 5) training samples. Specifically, from $\mathcal{D}_{unknown}$, small datasets (a.k.a episode/task) with individual training set and query set will be sampled. In each episode, the training set (a.k.a support set) contains $K$ classes $\{C^U_i\}_{i=1}^K \subset \mathcal{C}_{unknown}$  and $N$ samples $\{x^U_{ij}\}_{j=1}^N$ in each class $C^U_i$ (i.e. $K$-way $N$-shot), and the query set contains a query sample $x_q$ from $\{C^U_i\}_{i=1}^K$. The non-parametric testing on each novel-class episode is based on the Nearest-Neighbor classification. The probability that $x_q$ belongs to class $i$ is represented as

\vspace{-0.1cm}
\begin{equation}
P(y_i|x_q)=\frac{\exp(s(f_\theta(x_q), p^U_i))}{\sum_{k=1}^{K} \exp(s(f_\theta(x_q), p^U_k)) }
\label{eq:MN_P}
\end{equation}
where $f_\theta()$ is the network with parameter $\theta$, $p^U_i = \frac{1}{N} \sum_{j=1}^{N} f_\theta(x^U_{ij})$ is the prototype of the class $C^U_i$, and $s(,)$ is a similarity function (cosine similarity). The averaged performance on the episodes sampled from $\mathcal{D}_{unknown}$  will be the final performance of the model.

To enable the model of the novel-class Nearest-Neighbor classification, the known-class training must provide a feature extractor in the embedding space.
While training on $\mathcal{C}_{known}$, simply training a classifier (CNN backbone + fully connected (FC) classification layer) on it remains a strong baseline~\cite{li2019large, DBLP:journals/corr/abs-1904-04232, hariharan2017low, qiao2017few, rusu2019meta, tokmakov2019learning}. Combined with the cosine-similarity-based novel-class classification, it can be regarded as the baseline model for FSL. In this model, the backbone CNN's output $f_\theta(x) \in \mathbb{R}^{D \times 1}$ is regarded as the feature of the input $x$, and $D$ denotes the number of channels. The forward pass of the FC layer can be represented as $W^\top f_\theta(x)$, where $W\in \mathbb{R}^{D \times M}$ is the parameter of the FC layer, $M$ denotes the number of known classes, and we follow \cite{qiao2017few,gidaris2018dynamic} to discard the bias term. In the commonly used modified version for FSL, the feature and the FC parameters are $L_2$ normalized~ \cite{qiao2017few,gidaris2018dynamic}, we denote them as $f^c_\theta(x) = \frac{f_\theta(x)}{||f_\theta(x)||_2}$ and $W^c_{:,i} = \frac{W_{:,i}}{||W_{:,i}||_2}$. The classification loss is

\vspace{-0.1cm}
\begin{equation}
L_{classification} = -log( \frac{\exp(\tau {W^c_{:,y}}^\top f^c_\theta(x))}{\sum_{k=1}^{M} \exp(\tau {W^c_{:,k}}^\top f^c_\theta(x)) } )
\label{eq:cosine prob}
\end{equation}
where $y$ is the label of $x$, and $\tau$ is set to 30.0 following \cite{deng2019arcface} which controls the peakiness of the probability distribution.

\vspace{-0.1cm}
\subsection{Primitive discovery}

Visual primitives can refer to object parts~\cite{tokmakov2019learning}. Therefore, the primitive learning procedure should encourage the model to recognize object parts.
However, current methods~\cite{Vinyals2016Matching, snell2017prototypical} mainly rely on the image-level class labels for representation learning, which lacks the supervision for learning object parts. An alternative way is to manually label all possible object parts in each image. However, such annotation is prohibitively expensive and laborious. 
On the other hand, in the compositional recognition, transferring primitives learned on known classes to novel classes assumes they are semantically related. However, various classes with different semantic gaps may exist in novel classes.
To facilitate the discovery of part-related primitives without laborious annotations and reduce the effect of semantic gaps, we propose to use the self-supervision from object split orders to assist primitive learning.
Suppose we have an input image $x$ to be classified (shown in Fig.\ref{fig: framework}), splitting it into pieces will be likely to generate image patches containing different parts of the object in $x$. which encourages the model to recognize splits by recognizing parts, thus discovering primitives and latently encoding them in the network.
On the other hand, as objects with relatively large semantic gaps may share similar structures (e.g., the upper and lower parts of dogs and cars can be easily distinguished, although they are not highly semantically related), self-supervision from object structures~\cite{noroozi2016unsupervised, kim2019self} may help to alleviate the influence of semantic gaps.
Therefore, inspired by current works~\cite{gidaris2019boosting,kim2019self,mangla2019charting} on self-supervised learning, we propose to use the split-based self-supervised mechanism for FSL, where we split the input image horizontally and vertically, perm the splits, and ask the model to recognize which perms are applied to them. 

Specifically, given an input image $x$, we first divide it along rows and columns into $h \cdot v$ splits. The feature extractor will be applied to all splits to get $h \cdot v$ features, which will be denoted as $\{f_\theta(x_{rc})\}_{rc}^{h \cdot v}$, where $r$ and $c$ are the row id and column id respectively. We then randomly permute these splits to get a permuted sequence of splits, and ask the model to predict which permutation is applied to this sequence.
This permutation-prediction task is modelled as a classification problem. Then, the permuted features are concatenated and a FC layer is applied for classification. 
However, if the number of splits is large, there will be $(h \cdot v) !$ orders, and classifying the permuted sequence into all these classes will be difficult. To solve this problem, $M^s$ permutations with max Hamming distances~\cite{noroozi2016unsupervised} are used to permute the pieces. Denoting the concatenated feature vector as $f^s \in \mathbb{R}^{h \cdot v \cdot D}$, the loss is calculated as 

\vspace{-0.3cm}
\begin{equation}
L_{split} = -log(P(y^s|f^s))
\label{eq:jigsaw}
\end{equation}
where $y^s$ is the permutation label of $f^s$. 
Note that although the part-related primitives are strengthened, other kinds of primitives are not sacrificed, because the classification loss in equation~\ref{eq:cosine prob} still dominants the training process.


\vspace{-0.1cm}
\subsection{Primitive enhancing}

Assume we have well-trained primitives, the next step is to compose novel classes.
To achieve this goal, we must first know how primitives are represented in the deep network. Below we first provide our compositional view of the FSL baseline model with inspirations from current studies on the interpretability of deep networks, and then introduce our proposed composition mechanism.

\begin{figure}[t]
	\centering
	\includegraphics[width=0.65\columnwidth, height=0.27\columnwidth]{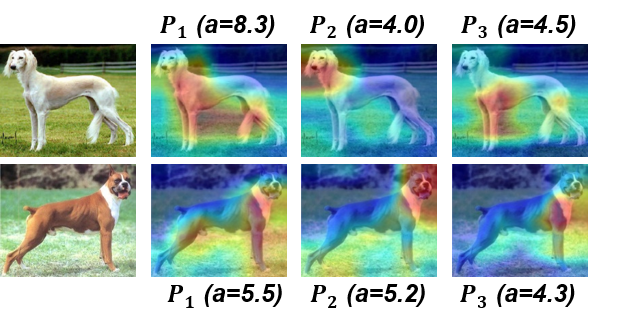}\vspace{-0.35cm}
	\caption{Primitives are shared as similar activated regions.}
	\label{fig: simple_share}\vspace{-0.4cm}
\end{figure}

Denote the feature map of $f^c_\theta(x)$ at channel $j$ as $A_\theta(x)_j \in \mathbb{R}^{h \times w}$. Given various input $x$, after mapping $A_\theta(x)_j$ to the original size of the input, current studies~\cite{bau2017network, zhou2016learning} on the interpretability of deep networks show that the heatmap of each channel can correspond to some certain patterns in the input. 
As shown in Fig.~\ref{fig: simple_share}, 
we can see that (a) given the same input, the feature maps of different channels have activation on different regions (Fig.~\ref{fig: simple_share} rows); (b) given different inputs, the same channel has activation on similar regions with different magnitude (Fig.~\ref{fig: simple_share} columns). For example, given images of dogs from two classes, channel 1, 2 and 3 have activation on similar parts around chest+fore$\&$rear legs, head, and chest+fore legs respectively with different magnitude.
The phenomenon (a) means the network is capable of learning various interested regions, and these regions can be viewed to compose the whole region that the network focuses on (e.g. CAM~\cite{zhou2016learning}).
The phenomenon (b) indicates that channels are transferable between different classes (at least to some degree).
Therefore, we view each channel of $f^c_\theta(x)$ as a visual primitive. The known-class classification probability of the input can be viewed as the normalized weighted sum of the activation on primitives, where the weights of primitives are represented in the FC layer's parameters.
Then, during the novel-class testing, as all channels are already contained in the extracted novel-class feature, the Nearest Neighbor classification on novel classes can be viewed as the comparison of two primitive sets,
because the calculation of the cosine similarity (i.e., $L_2$ normalized dot product) can be viewed as comparing the similarity on the activation of each primitive respectively, and outputting an overall similarity score.

However, as primitives are equally weighted in the novel-class feature of the baseline model, the composition may be harmed by influences from those trivial primitives.
One straightforward solution may be the hard composition mechanism, which explicitly selects all important primitives and neglect those trivial ones.
However, to achieve the hard composition, we must modify the network structure to allow a dynamic number of all possible primitives, which is complicated and difficult to implement.
Therefore, to achieve the effective composition and simplify the hard composition, we propose a soft composition mechanism by an Enlarging-Reducing loss (ER loss) to enhance the influence of those important primitives while reducing that of others, making those important primitives nearly influence all the novel-class classification, so that the novel classes can be approximately viewed as being composed by these primitives. Compared with the hard composition mechanism, the soft composition mechanism keeps a fixed set of all candidate primitives and requires no extra modification in the baseline network structure.
Moreover, combining the compositional view, we can then explain which primitives from known classes compose the given novel sample by visualizing the feature maps, which shows better interpretability for the deep-learning-based FSL (see section~\ref{sec: Experiments: visualization}).

The ER loss is inspired by both mathematically and the biologically. As both inspirations give the same formulation of the adopted loss, below we first give the deduction from the mathematical view, and then the inspiration from biological studies will be included.

\subsubsection{Inspiration from mathematical deduction}\qquad

In the compositional view of the FSL baseline model, primitives are transferable across classes although classes are not overlapping.
As the importance of each primitive to class $i$ is represented in $W^c_{:, i}$,
we can select important primitives for each known class from the total primitive set, and the selected ones intuitively should also have higher influence in the novel-class classification.

In FSL~\cite{Vinyals2016Matching}, the cosine similarity function is widely adopted as the similarity function $s(,)$ in equation~\ref{eq:MN_P}. 
For simplicity, we handle the situation where the number of shot is 1 (i.e., $p^U_i = f_\theta(x^U_{i})$). 
As shown in equation~\ref{eq:MN_P}, now the similarity is computed as

\vspace{-0.3cm}
\begin{equation}
s(f_\theta(x_q), f_\theta(x^U_i)) = \sum_{j=1}^{D} {f^c_\theta(x_q)}_j \cdot {f^c_\theta(x^U_i)}_j
\label{eq:cosine cal}
\end{equation}
Now we define the influence of the primitive represented by ${f^c_\theta()}_k$ in novel-class classification as 

\vspace{-0.3cm}
\begin{align}
	influ^U = \frac{{f^c_\theta(x_q)}_k \cdot {f^c_\theta(x^U_i)}_k}{s(f_\theta(x_q), f_\theta(x^U_i))} = \frac{{f^c_\theta(x_q)}_k \cdot {f^c_\theta(x^U_i)}_k}{\sum_{j=1}^{D} {f^c_\theta(x_q)}_j \cdot {f^c_\theta(x^U_i)}_j} \nonumber\\
	= \frac{1}{(\sum_{j=1,j \neq k}^{D} {f^c_\theta(x_q)}_j \cdot {f^c_\theta(x^U_i)}_j) / ({f^c_\theta(x_q)}_k \cdot {f^c_\theta(x^U_i)}_k) + 1}
	\label{eq:importance of pattern in uknown class}
\end{align}

To increase the influence of the primitive represented by ${f^c_\theta()}_k$, there are two ways:

a) Enlarging the term ${f^c_\theta(x_q)}_k \cdot {f^c_\theta(x^U_i)}_k$

b) Reducing the term $\sum_{j=1,j \neq k}^{D} {f^c_\theta(x_q)}_j \cdot {f^c_\theta(x^U_i)}_j$

As primitives are shared between known classes and novel classes, we can simply achieve this goal in known-class training. To better compose novel classes with learned primitives, we propose the Enlarging-Reducing (ER) loss in known-class training as follows,

\vspace{-0.35cm}
\begin{equation}
L_{ER} = - \lambda_1 \sum_{j \in T(W_{:,y}, D^*)} f_\theta(x)_j + \lambda_2 \sum_{j \notin T(W_{:,y}, D^*)} f_\theta(x)_j
\label{eq:enlarge-reduce loss}
\end{equation}
where $y$ is the class that the known-class sample $x$ belongs to, and $T(, D^*)$ returns the indices of top $D^*$ elements in the given vector. Top $D^*$ elements in $W_{:,y}$ represent the primitives that contribute the most to the classification of $x$ to its class $y$. 
In the novel-class classification, if the primitives from $T(W_{:,y}, D^*)$ influences nearly all the similarity calculation, the novel classes can be viewed to be composed by these primitives. Relevant experiments are in Fig.~\ref{fig: topK}.




Combine the primitive discovery, the total training loss is 

\vspace{-0.4cm}
\begin{align}
	L = L_{classification} + \alpha_1 L_{split} + \alpha_2 L_{ER}
	\label{eq:dinal loss}
\end{align}
where $\alpha_1, \alpha_2$ are pre-defined hyper-parameters. After training on known classes, the trained network will be used to perform Nearest-Neighbor classification on novel classes as stated in section~\ref{sec: preliminaries}.

\vspace{-0.1cm}
\subsubsection{Inspiration from biological studies}\qquad
\label{sec:bio_ER}

\textbf{Hebbian Learning} is a widely observed unsupervised learning mechanism in human brain~\cite{hebb1949organization} related to memory. 
It is stated as when an axon of cell A is near enough to excite a cell B and repeatedly or persistently takes part in firing it, some growth process or metabolic change takes place in one or both cells such that A's efficiency, as one of the cells firing B, is increased~\cite{hebb1949organization}.
In known-class classification, the forward pass of the FC layer (i.e. ${W^c_{:,i}}^\top f^c_\theta(x)$) can be viewd as connected cells. Each connection is between $W^c_{ji}$ and $f^c_\theta(x)_j$. 
In Fig.~\ref{fig: distribution} we observe that large $f^c_\theta(x)_j$ always has large ${W_{jy}^c}$ ($y$ stands for the label of the input $x$) multiplied with it. 
As the firing of a neural cell is always observed when its membrane voltage is higher than a threshold~\cite{glorot2011deep}, we can view ${W^c_{jy}}$ and $f^c_\theta(x)_j$ as firing cells when their values are large. As large $f^c_\theta(x)_j$ is always connected with large ${W^c_{jy}}$, cells represented by them always fire together. 
As ${W^c_{jy}}$ is easier to have a relatively larger value due to the easiness in the back propagation, 
according to Hebbian Learning, 
we increase the efficiency of $f^c_\theta(x)_j$ by enlarging its value.
This mechanism can be represented by the first term in the ER loss.


\textbf{Winner-Take-All mechanism} is widely used in the brain cortex learning and the Spiking Neural Network~\cite{douglas2004neuronal}. It means that columns in the cortex are competing against each other, and the winner will suppress others, producing sparse spikes. While the feature extractor is being trained on known classes, each feature channel $f_\theta()_j$ can be viewed as competing against each other to get larger importance in the classification, where its importance to class $y$ is measured by $W^c_{jy}$.
According to the Winner-Take-All mechanism, the winner should suppress all other losers, and this can be represented by the second term in the ER loss. 


\vspace{-0.2cm}
\subsection{Auxiliary objective and regularization}

As we view $W$ as the weights for the corresponding primitive, when $W$ is smaller, the importance of the primitive should also be lower. However, the activation of primitives can be negative (e.g., LSTM~\cite{Donahue2015Long}), if $W$ is also negative, the dot product will be positive, disobeying what we want. So we use $W = abs(W)$ to avoid such situation. As the weights in $W$ are continuous and it is hard to draw a clear line between high and low, we add a sparseness loss on each column of $W$, which can be represented as $\frac{1}{M} \sum_{i=1}^{M} ||W_{:,i}||_1$, which constrains the model to use few primitives in known-class recognition and improve the quality of each primitive, and making it easier to control the ER loss hyper-parameters. The weight for this loss is typically 0.1. To learn the object structure implicitly, we also add the self-supervised rotation loss~\cite{mangla2019charting, gidaris2019boosting} to assist the learning. Given an image, it is rotated by \{0, 90, 180, 270\} degrees, and the model is asked to predict which rotation is applied to the image. This self-supervised loss can also help alleviating the influence from semantic gaps.
Compared with this term, our model explicitly learn primitives related to object parts by splitting images into splits.

\vspace{-0.2cm}
\section{Experiments}
To verify the proposed methods, we conduct extensive experiments on both few-shot image and video classification. We first introduce the datasets and implementation details. Then the comparison with state-of-the-art and the ablation study will be reported. We also visualize the primitives learned by our method, so as to provide interpretability of the proposed method. Due to the space limitation, please refer to the supplementary material for more details.


\textbf{Datasets and settings.} 
Experiments on few-shot image classification are conducted on the CUB-200-2011 (CUB)~\cite{wah2011caltech} and the \textit{mini}ImageNet~\cite{Vinyals2016Matching} benchmarks, while experiments on few-shot video classification are conducted on the Kinetics dataset~\cite{Kay2017The}.
CUB contains 200 fine-grained bird classes and 11,788 images in total. Following the settings in \cite{DBLP:journals/corr/abs-1904-04232, qiao2019transductive}, we split the dataset into 100 training  classes, 50 validation classes and 50 test classes.
\textit{Mini}ImageNet has 100 classes selected from the ImageNet~\cite{deng2009imagenet} with 600 images in each class, which has 64 training classes, 16 validation classes and 20 test classes.
Kinetics is introduced into the few-shot video recognition task by CMN~\cite{zhu2018compound}. Following the provided splits, it contains 100 classes of actions with 100 videos in each class in total. We follow CMN to split the dataset into 64 training classes, 12 validation classes and 24 test classes. Following existing methods~\cite{Vinyals2016Matching,zhu2018compound}, the mean accuracy (\%) and the 95\% confidence intervals of randomly generated episodes on the test (novel) sets will be reported.

\begin{table}[t]
	\footnotesize
	\begin{center}
		\caption{Evaluative results(\%) on CUB.}\vspace{-0.3cm}
		\label{tab:CUB}
		\begin{tabular}{l|c|c|c}
			\hline\hline
			Method  & Backbone & 5-way 1-shot & 5-way 5-shot
			\\
			\hline
			MatchingNet~\cite{Vinyals2016Matching} & Conv4 & $61.16 \pm 0.89$ & $72.86 \pm 0.70$ \\
			ProtoNet~\cite{snell2017prototypical} & Conv4 & $51.31 \pm 0.91$ & $70.77 \pm 0.69$ \\
			MAML~\cite{finn2017model} & Conv4 & $55.92 \pm 0.95$ & $72.09 \pm 0.76$ \\
			RelationNet~\cite{yang2018learning} & Conv4 & $62.45 \pm 0.98$ & $76.11 \pm 0.69$ \\
			DEML+MetaSGD~\cite{zhou2018deep} & ResNet50 & $66.95 \pm 1.06$ & $77.1 \pm 0.78$ \\
			ResNet18+TriNet~\cite{chen2018semantic} & ResNet18 & $69.61 \pm 0.46$ & $84.10 \pm 0.35$ \\
			MAML++~\cite{antoniou2019learning} & DenseNet & $67.48 \pm 1.44$ & $83.80 \pm 0.35$ \\
			SCA + MAML++~\cite{antoniou2019learning} & DenseNet & $70.33 \pm 0.78$ & $85.47 \pm 0.40$ \\
			S2M2~\cite{mangla2019charting} & ResNet18 & $72.40 \pm 0.34$ & $86.22 \pm 0.53$ \\
			CFA~\cite{hu2019weakly} & ResNet18 & $73.90 \pm 0.80$ & $86.80 \pm 0.50$ \\
			\hline
			Cosine Classifier & ResNet18 & $72.22 \pm 0.33$ & $86.41 \pm 0.18$ \\
			CPDE (ours) & ResNet18 & $\textbf{80.11} \pm \textbf{0.34}$ & $ \textbf{89.28} \pm \textbf{0.33}$ \\
			\hline\hline
		\end{tabular}
	\end{center}\vspace{-0.5cm}
\end{table}

\textbf{Implementation details.}
Our model is implemented with the TensorFlow~\cite{abadi2016tensorflow}. 
The Nesterov Momentum optimizer~\cite{sutskever2013importance} is used with an initial learning rate of 0.01. The total training epochs on the CUB, the \textit{mini}ImageNet and the Kinetics are 57, 40 and 23, and the learning rate is dropped to 10\% on (30, 40), (30, 37), (2, 19) epochs respectively. 
The weight decay is set to be 0.0005. 
Typical data augmentation methods such as random flipping and random brightness are adopted.
In $L_{split}$, the rows and columns are set to 2, the weight $\alpha_1$ is set to 0.5 for all datasets. In $L_{ER}$, the weight $\lambda_1$ is set to 1.0 and $\lambda_2$ is set to 0.5, and $D^*$ is set to 5. The overall weight $\alpha_2$ is set to 0.1. Hyper-parameters are chosen on the validation set.

\vspace{-0.1cm}
\subsection{Comparison with state-of-the-art}
\label{sec: Experiments: comparison}

\begin{table}[t]
	\footnotesize
	\begin{center}
		\caption{Evaluative results(\%) on the \textit{mini}ImageNet.}\vspace{-0.3cm}
		\label{tab:miniImagenet}
		\begin{tabular}{l|c|c|c}
			\hline\hline
			Method  & Backbone & 5-way 1-shot & 5-way 5-shot
			\\
			\hline
			MatchingNet~\cite{Vinyals2016Matching} & Conv4 & $46.56 \pm 0.84$ & $55.31 \pm 0.73$ \\
			ProtoNet~\cite{snell2017prototypical} & Conv4 & $49.42 \pm 0.78$ & $68.20 \pm 0.66$ \\
			MAML~\cite{finn2017model} & Conv4 & $48.70 \pm 1.84$ & $63.11 \pm 0.92$ \\
			RelationNet~\cite{yang2018learning} & Conv4 & $50.44 \pm 0.82$ & $65.32 \pm 0.70$ \\
			AgileNet~\cite{ghasemzadeh2018agilenet} & Conv4 & $58.23 \pm 0.10$ & $71.39 \pm 0.10$ \\
			DEML+MetaSGD~\cite{zhou2018deep} & ResNet50 & $58.49 \pm 0.91$ & $71.28 \pm 0.69$ \\
			Dynamic FS~\cite{gidaris2018dynamic} & ResNet10 & $55.45 \pm 0.89$ & $70.13 \pm 0.68$ \\
			SNAIL~\cite{mishra2017simple} & ResNet12 & $55.71 \pm 0.99$ & $ 68.88 \pm 0.92 $ \\
			TADAM~\cite{oreshkin2018tadam} & ResNet12 & $58.50 \pm 0.30$ & $76.70 \pm 0.30$ \\
			Acti (\textit{trainval})~\cite{qiao2017few} & WRN-28-10 & $59.60 \pm 0.41$ & $73.74 \pm 0.19$ \\
			LEO (\textit{trainval})~\cite{rusu2019meta} & WRN-28-10 & $ 61.76 \pm 0.08 $ & $ 77.59 \pm 0.12 $ \\
			DCO~\cite{lee2019meta} & ResNet12 & $62.62 \pm 0.61 $ & $ 78.63 \pm 0.46$ \\
			DCO (\textit{trainval})~\cite{lee2019meta} & ResNet12 & $64.09 \pm 0.62 $ & $ 80.00 \pm 0.45$ \\
			CTM~\cite{li2019finding} & ResNet18 & $ 64.12 \pm 0.82 $ & $ 80.51 \pm 0.13 $ \\
			\hline
			Cosine Classifier & ResNet10 & $55.97 \pm 0.26$ & $ 74.95 \pm 0.24 $ \\
			Ours & ResNet10 & $ 62.66 \pm 0.69$ & $ 77.45 \pm 0.71 $ \\
			Ours (\textit{trainval}) & ResNet10 & $ \textbf{64.37} \pm \textbf{0.77} $ & $ \textbf{79.10} \pm \textbf{0.74} $ \\
			\hline
			Cosine Classifier & ResNet12 & $56.26 \pm 0.28$ & $ 74.97 \pm 0.24 $ \\
			Ours & ResNet12 & $ 63.21 \pm 0.78 $ & $ 79.68 \pm 0.82 $ \\
			Ours (\textit{trainval}) & ResNet12 & $ \textbf{64.17} \pm \textbf{0.84} $ & $ \textbf{80.47} \pm \textbf{0.89} $ \\
			\hline
			Cosine Classifier & ResNet18 & $56.92 \pm 0.28$ & $ 75.39 \pm 0.24 $ \\
			Ours & ResNet18 & $ 64.44 \pm 0.79 $ & $ 79.06 \pm 0.57 $ \\
			Ours (\textit{trainval}) & ResNet18 & $ \textbf{65.55} \pm \textbf{0.72} $ & $ \textbf{80.66} \pm \textbf{0.75} $ \\
			\hline\hline
		\end{tabular}
	\end{center}\vspace{-0.5cm}
\end{table}

\begin{table}[t]
	\footnotesize
	\begin{center}
		\caption{Evaluation(\%) on 5-way few-shot action recognition.}\vspace{-0.3cm} 
		\label{tab:Kinetics}
		\begin{tabular}{l|c|c}
			\hline\hline
			Method  & \tabincell{c}{5-way 1-shot} & \tabincell{c}{5-way 5-shot}
			\\
			\hline
			RGB w/o mem  & $28.7$ & $48.6$ \\
			Flow w/o mem  & $24.4$ & $33.1$ \\
			LSTM(RGB) w/o mem  & $28.9$ & $49.0$ \\
			Nearest-finetune  & $48.2$ & $62.6$ \\
			Nearest-pretrain & $51.1$ & $68.9$ \\
			Matching Network~\cite{Vinyals2016Matching}  & $53.3$ & $74.6$ \\
			MAML~\cite{finn2017model}  & $54.2$ & $75.3$ \\
			Plain CMN~\cite{kaiser2017learning}  & $57.3$ & $76.0$ \\
			LSTM-cmb & $57.6$ & $76.2$ \\
			CMN~\cite{zhu2018compound} & $60.5$ & $78.9$ \\
			TARN~\cite{bishay2019tarn} & $66.55$ & $80.66$ \\
			\hline
			Cosine classifier & $67.05 \pm 0.72$  & $80.00 \pm 0.59$ \\
			CPDE (ours) & $\textbf{69.14} \pm \textbf{0.68}$ & $ \textbf{82.19} \pm \textbf{0.60} $ \\
			\hline\hline
		\end{tabular}
	\end{center}\vspace{-0.48cm}
\end{table}

\begin{table}[]
	\footnotesize
	\begin{center}
		\caption{Top-5 accuracy(\%) for the 100-way classification task on CUB. Note that in CompCos, attribute annotations are used. We can achieve comparable (even slightly better in 1-shot) performance \textit{without such annotations}.}\vspace{-0.3cm}
		\label{tab:CUB_yuxiong}
		\begin{tabular}{l|c|c|c}
			\hline\hline
			Method & 100-way 1-shot & 100-way 2-shot & 100-way 5-shot
			\\
			\hline
			ProtoNet~\cite{snell2017prototypical} & $43.2$ & $54.3$ & $67.8$ \\
			MatchingNet~\cite{Vinyals2016Matching} & $48.5$ & $57.3$ & $69.2$ \\
			RelationNet~\cite{yang2018learning} & $39.5$ & $54.1$ & $67.1$ \\
			CompCos~\cite{tokmakov2019learning} & $53.6$ & $\textbf{64.8}$ & $\textbf{74.6}$ \\
			\hline
			Cosine Classifier & $ 48.22 \pm 0.38 $ & $ 60.79 \pm 0.43 $ & $ 72.81 \pm 0.39 $ \\
			CPDE (ours) & $ \textbf{54.01} \pm \textbf{0.56} $ & $ 64.73 \pm 0.51 $ & $ 74.14 \pm 0.41 $ \\
			\hline\hline
		\end{tabular}
	\end{center}\vspace{-0.5cm}
\end{table}

Comparative evaluation with existing algorithms on few-shot image recognition can be illustrated in Tables~\ref{tab:CUB}, ~\ref{tab:miniImagenet}, ~\ref{tab:Kinetics} and ~\ref{tab:CUB_yuxiong}. We denote our baseline model as the Cosine Classifier.
On the CUB dataset, we follow the same data pre-processing as \cite{ye2018learning,triantafillou2017few, qiao2019transductive} to use the available bounding boxes to crop the images.
On the \textit{mini}ImageNet, we follow \cite{qiao2017few,rusu2019meta,lee2019meta} to train the model using the training set with (\textit{trainval}) and without the validation set.
In Tables~\ref{tab:CUB} and~\ref{tab:miniImagenet}, our method can achieve significantly better performance than existing methods in both 5-way 1-shot and 5-way 5-shot tasks.
For few-shot action classification, we follow~\cite{zhu2018compound} to adopt the ResNet50~\cite{He_2016_CVPR} as our backbone and pre-train it on the ImageNet. 
To capture the temporal information, we add a 1-layer LSTM~\cite{Donahue2015Long} with 512 units on the top of the ResNet50, and the split loss and the rotation loss are not applied.
We averagely sample 5 RGB frames from each video.
Similarly, superior performance of our method to the state-of-the-art methods can be observed in Table \ref{tab:Kinetics}.
To compare with the compositional method CompCos~\cite{tokmakov2019learning} fairly, we follow them to use the ResNet10~\cite{He_2016_CVPR} backbone, and split the CUB dataset to 100 known classes and 100 novel classes. The evaluation are carried on the 100-way classification of the novel classes in Table~\ref{tab:CUB_yuxiong}. Note that in CompCos, attribute annotations are used, while in our method, \textbf{we do not use such annotations} but achieve comparable  performance (even slightly better in 1-shot).

\vspace{-0.1cm}
\subsection{Ablation study} \label{sec: Experiments: ablation study}

\subsubsection{Verification of each module}

The ablation study on the effect of different modules in (\ref{eq:dinal loss}) are shown in Table~\ref{tab:CUB_ablation}. 
The model in each line on the left hand side consists of the modules all the above and the one listed in current line. 
As shown in Table~\ref{tab:CUB_ablation}, each module is verified its rationale and positive effect on improving classification performance, especially the ER loss in the 1-shot scenario.

\begin{table}[t]
	\footnotesize
	\begin{center}
		\caption{Ablation study of each module. }\vspace{-0.3cm}
		\label{tab:CUB_ablation}
		\begin{tabular}{c|c|c|c}
			\hline\hline
			Study Case & Method  & 5-way 1-shot (\%) & 5-way 5-shot (\%)
			\\
			\hline
			\multirow{4}{*}{\tabincell{c}{CUB \\ ResNet18}} & cosine classifier & $ 72.22 \pm 0.33 $ & $ 86.41 \pm 0.18 $ \\
			& + auxiliary terms & $ 74.14 \pm 0.33 $ & $ 87.49 \pm 0.48 $\\
			& + $L_{split}$ & $ 76.10 \pm 0.37 $ & $ 88.09 \pm 0.24 $\\
			& + $L_{ER}$ & $\textbf{80.11} \pm \textbf{0.34}$ & $ \textbf{89.28} \pm \textbf{0.33} $ \\
			\hline
			\multirow{4}{*}{\tabincell{c}{\textit{mini}ImageNet \\ ResNet10 }} & cosine classifier & $ 55.97 \pm 0.26 $ & $ 74.95 \pm 0.24 $ \\
			& + auxiliary terms & $ 58.35 \pm 0.26 $ & $  76.16 \pm 0.56 $\\
			& + $L_{split}$ & $ 59.96 \pm 0.29 $ & $  76.51 \pm 0.25 $\\
			& + $L_{ER}$ & $\textbf{62.66} \pm \textbf{0.69}$ & $ \textbf{77.45} \pm \textbf{0.71} $ \\
			\hline
			\multirow{3}{*}{\tabincell{c}{Kinetics \\ ResNet50 }} & cosine classifier & $ 67.05 \pm 0.72 $ & $ 80.00 \pm 0.59 $ \\
			& + auxiliary terms & $ 67.84 \pm 0.53 $ & $   80.68 \pm 0.61 $\\
			& + $L_{ER}$ & $\textbf{69.14} \pm \textbf{0.68}$ & $ \textbf{82.19} \pm \textbf{0.60} $ \\
			\hline\hline
		\end{tabular}
	\end{center}\vspace{-0.6cm}
\end{table}

\subsubsection{Evaluation of primitive discovery}

\begin{figure}[t]
	\centering\includegraphics[width=1.0\columnwidth, height=0.4\columnwidth]{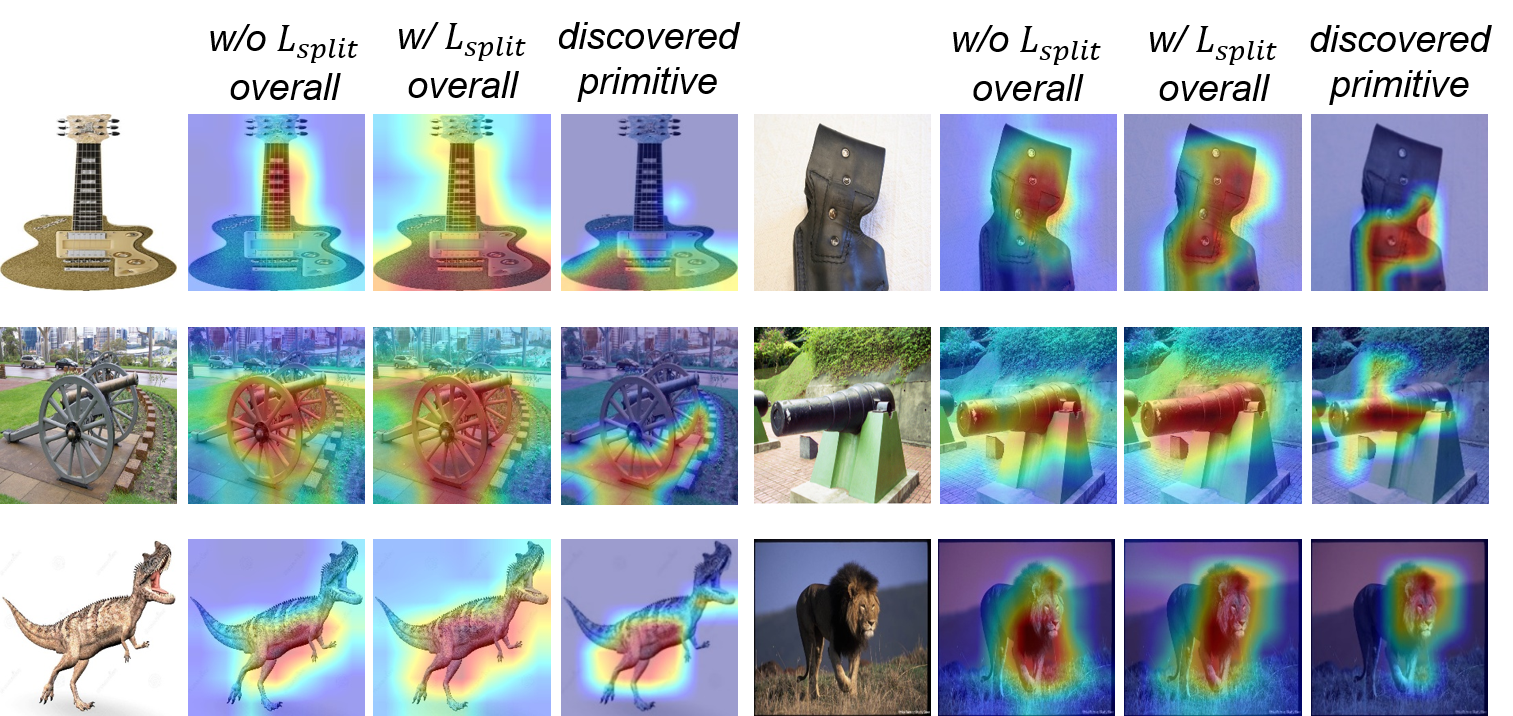}\vspace{-0.1cm}
	\caption{Heatmaps with(w/) and without(w/o) $L_{split}$. Images are from novel classes of \textit{mini}ImageNet. By adding $L_{split}$, the model can discover more activated regions. The discovered primitive is visualized in the last column, which refers to the single channel that contains the discovered region.}
	\label{fig: discovered primitives}\vspace{-0.4cm}
\end{figure}

To verify that the supervision from split orders can discover the part-related primitives, we visualized the heatmap of novel-class samples in Fig.~\ref{fig: discovered primitives}. We visualize them by calculating the weighted sum of the heatmap by $\sum_{j=1}^{D} f(x)_j \cdot A(x)_j$ (denoted as \textit{overall}), where all notations are the same as in the methodology.
We can find that compared with the model without $L_{split}$, adding it can have activation on more object parts. 
As the primitive is encoded in each channel of the penultimate layer of deep networks, we also visualize those single channels that cover the discovered regions.
This result can verify that the model is pushed to extract useful information from each split, leading to the discovery of part-related primitives. 

\vspace{-0.1cm}
\subsubsection{Evaluation of primitive enhancing}

\begin{figure}[t]
	\centering
	\subfigure{\includegraphics[width=0.36\columnwidth]{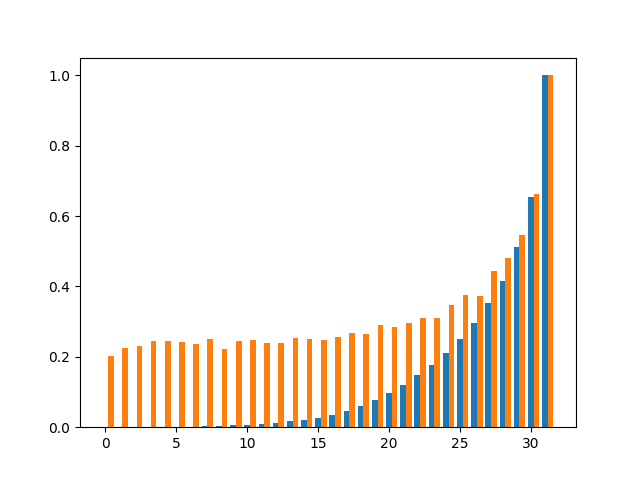}} \hspace{-5mm}
	\subfigure{\includegraphics[width=0.36\columnwidth]{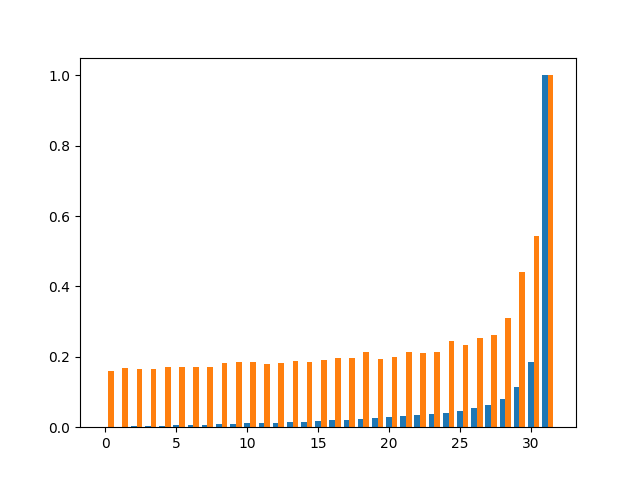}} \hspace{-5mm}
	\subfigure{\includegraphics[width=0.36\columnwidth]{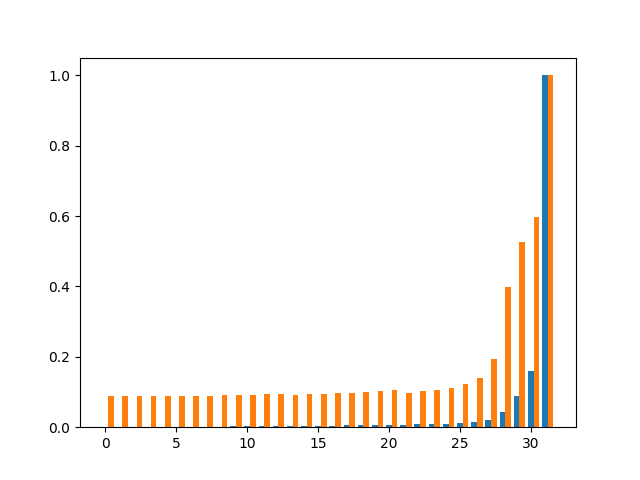}}\vspace{-0.3cm}
	\caption{Ablation study on primitive enhancing. Blue and orange bars represent the distribution of $W_{:,y}$ and distribution of primitive activation, which are both sorted according to an ascending order of the values in $W_{:,y}$. Left: Cosine Classifier +abs($W$). Middle: + sparseness loss. Right: + ER loss.
	}
	\label{fig: distribution}\vspace{-0.5cm}
\end{figure}

To study the effect of ER loss (\ref{eq:enlarge-reduce loss}) in primitive enhancing, we plot the distribution of weights and activation of all primitives in Fig.~\ref{fig: distribution}. Experiments are conducted with the ResNet10 trained on  \textit{mini}ImageNet. In Fig.~\ref{fig: distribution}, $W$ is depicted with blue bars, while orange bars represent the activation on primitives.
Given an input $x$, we first sort each channel in $W_{:,y}$ in the ascending order ($y$ denotes the corresponding class of $x$), whose indices are used to sort channels in $f_\theta(x)$.
We divide all channels into 32 bins (e.g. given a feature of 512 channels, there will be 16 channels in each bin), and calculate the average value of each bins to plot. All bins are divided by the max value among them for normalization.
We randomly select 1000 samples from known classes and calculate the average bins.
From Fig.~\ref{fig: distribution}, we can find that large $f_\theta()_j$ is usually accompanied with large $W_{jy}$ in the same feature channel, which verifies that $f^c_\theta(x)_j$ and ${W^c_{jy}}$ can be viewed as connected cells firing together.
By applying the auxiliary sparseness loss, in Fig.~\ref{fig: distribution}(middle), most weight dimensions are suppressed with relatively low values and sparse dimensions can have high response, making it easier to select the important primitives to enhance.
By applying the ER loss, activation of most primitives are enforced to have lower values in Fig.~\ref{fig: distribution}(right) compared with Fig.~\ref{fig: distribution}(middle), which means enhanced influence of those important primitives achieved by the ER loss. And it is consistent with the WTA inspiration of our model.

To evaluate the effect of important primitives, we first evaluate the novel-class classification performance $Acc_k$ of our model with top $k$ feature channels selected by their activation (i.e. set the values in other channels to 0), whose results are shown in Fig.~\ref{fig: topK}(left). 
Y-axis indicates the portion (\%) of $Acc_k$ credited to the performance when using all channels, i.e., $Acc_k / Acc_{all}$, which implies the contribution of selected primitives to the novel-class classification. 
As large activation tends to have large weights as shown in Fig.~\ref{fig: distribution}, remained primitives are more important.
Compared with the baseline cosine classifier depicted with the blue curve, applying the ER loss makes important primitives have larger influence, which verifies that the novel class is better composed by important primitives. 

Similarly, we then use top $k$ large values in each column of FC parameters $W$ to evaluate known-class classification (i.e. set values of other channels to 0 in each $W_{:, i}$), whose results are shown in Fig.~\ref{fig: topK}(right). 
Similarly, important primitives have larger influence in our method. Moreover, it also verifies the view that $W$ can be treated as measuring the importance of the corresponding primitive, because only few primitives selected by $W$ can indeed recover the total performance.
Note that due to the limited data, novel classes always need more primitives to represent, thus the number of important primitives is larger in novel-class classification.


\vspace{-0.1cm}
\subsection{Composition of primitives} \label{sec: Experiments: visualization}

\begin{figure}[t]
	\centering
	\subfigure{\includegraphics[width=0.48\columnwidth]{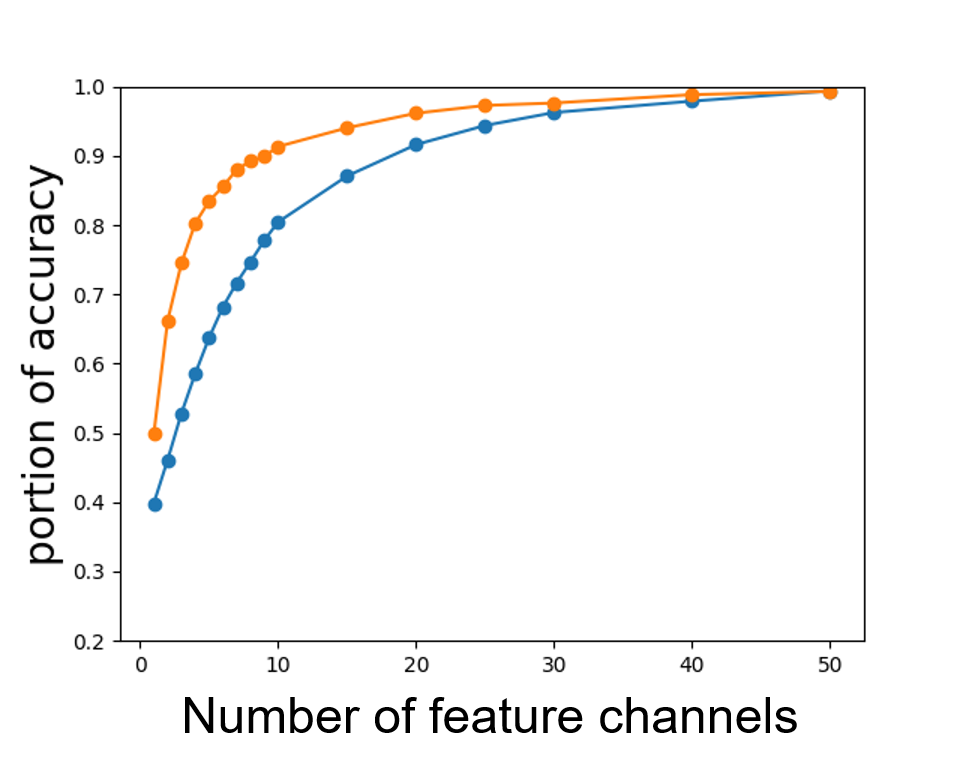}}
	\subfigure{\includegraphics[width=0.48\columnwidth]{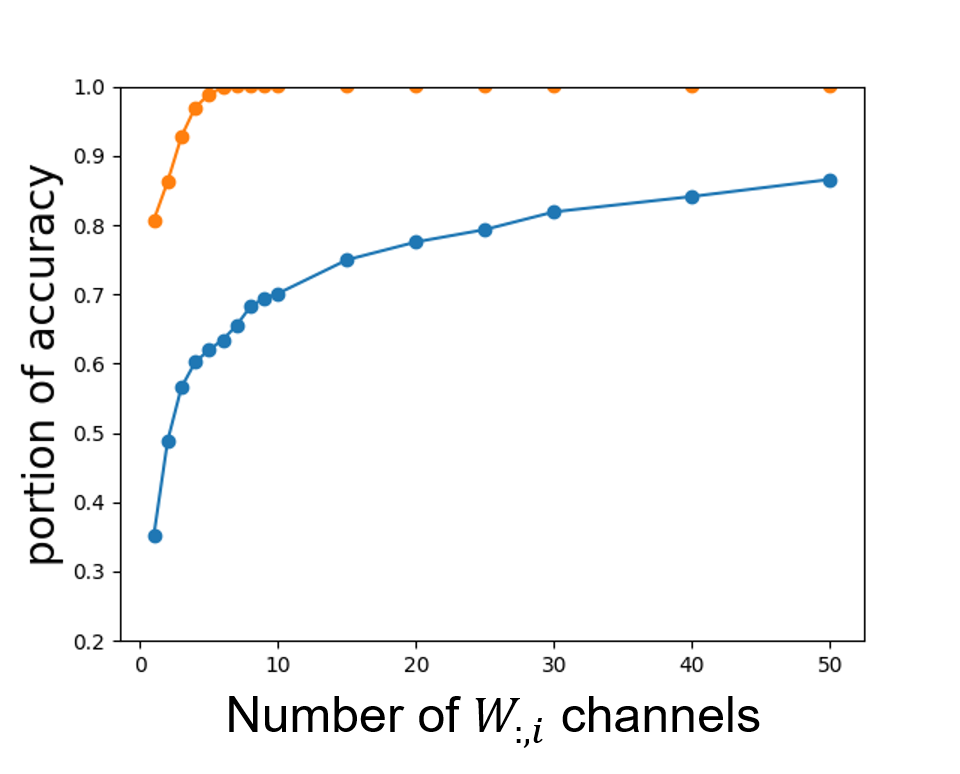}}\vspace{-0.3cm}
	\caption{Left: Portion of novel-class classification accuracy maintained v.s. number of top $k$ feature channels remained. Right: Portion of known-class classification accuracy maintained v.s. number of top $k$ elements in $W_{:,i}$ channels remained. 
		Our method is in orange, while the cosine classifier in blue. We can see that $W$ can select the most important primitives, and they have larger influence in our model.}
	\label{fig: topK}\vspace{-0.3cm}
\end{figure}

\begin{figure}[t]
	\centering\includegraphics[width=1.0\columnwidth, height=1.2\columnwidth]{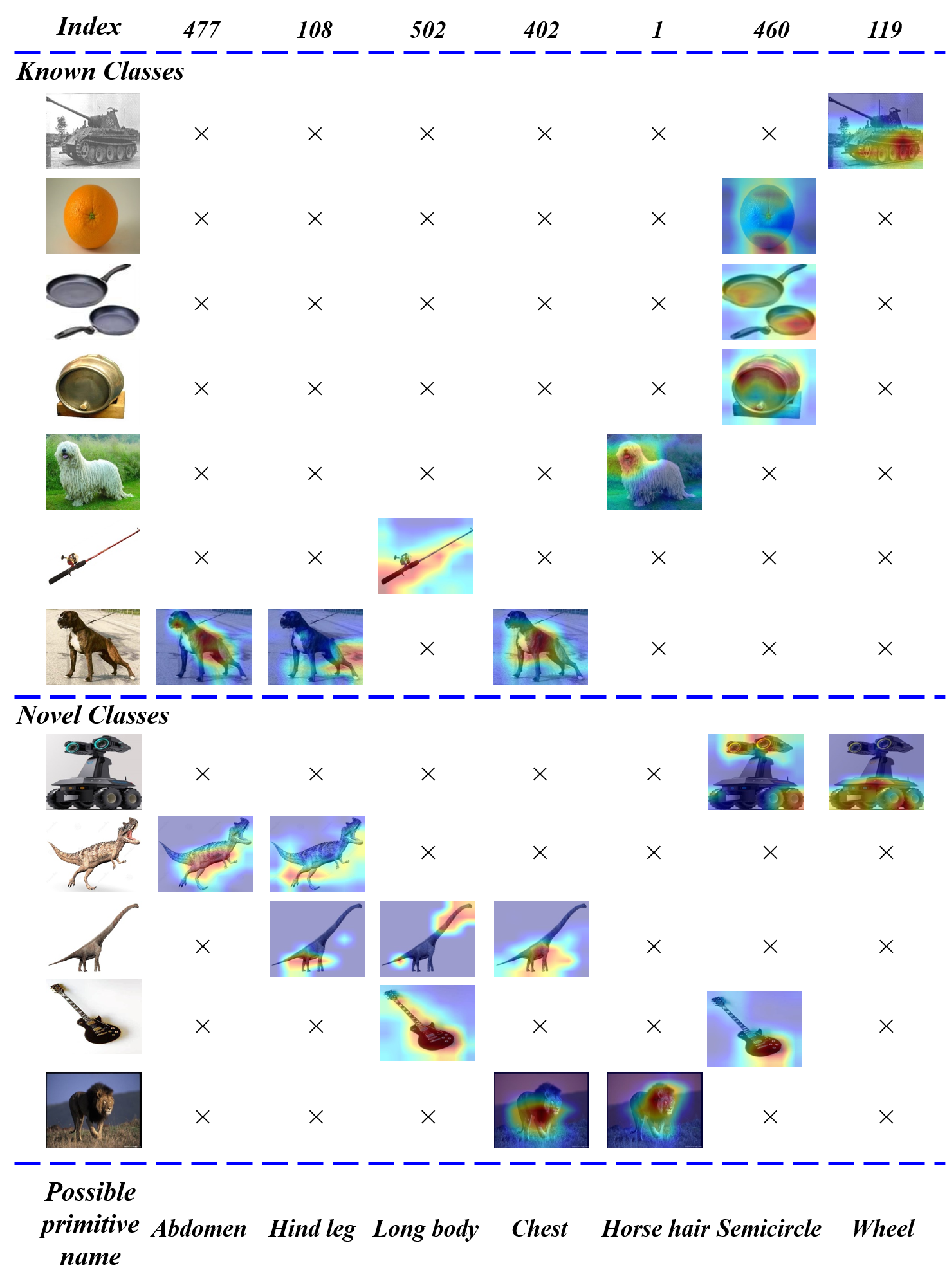}\vspace{-0.3cm}
	\caption{Visualization of primitives. Primitives in the same column share the channel index. Novel samples are composed of primitives learned on known classes. Most dimensions have no distinguishable activation ($\times$) due to the sparseness of features. }
	\label{fig: visualization}\vspace{-0.5cm}
\end{figure}

To show how novel classes are composed by primitives learned on known classes, we visualize primitives overlapped in known-class and novel novel-class classification in Fig.~\ref{fig: visualization}. 

In Fig.~\ref{fig: topK}(left), about 15 feature channels can recover 95\% of classification performance, so we choose the top 15 feature channels to mark down their indices as $ T(f_\theta(X^U), 15) $, where $T(a, K)$ means the set formed by top $K$ elements' indices of the vector $a$.
In Fig.~\ref{fig: topK}(right), only 5 channels can recover the total performance. 
Thus we mark down their indices as 
$ T(W_{:,i}, 5) $ for all known class $i$. 
We visualize the primitives overlapped in these two set, so that we can known which primitives in known classes compose the given novel class sample. 
However, as not all primitives are easy to understand, we only visualize those easy ones (most are related to object parts). As we don't have the manual annotated meaning of them, we write the possible meaning of them in the last row. The primitive/channel indices are written in the first row.
$\times$ denotes no distinguishable activation in that channel (i.e. not in $ T(W_{:,i}, 5) $ or $ T(f_\theta(X^U), 15) $). 
Most channels have low responses due to the sparseness of features.
Known classes and last two rows of novel classes are from the \textit{mini}ImageNet. 
We can find that primitives in the same feature channels have high response in similar spatial regions of object parts, which are marked as red. 
With this visualization, for example, we can explain the dinosaur in the third row of novel classes as \textit{having the legs and chest structure like a dog from the side view and having a neck like a fishing rod}, and explain the lion in the last row as \textit{having chest like a dog and having horse hair like a hairy dog}.

\vspace{-0.1cm}
\section{Conclusion}
We propose a novel FSL approach to imitate humans to recognize novel classes by composing primitives learned from known classes. Our method utilizes self-supervision to discover part-related primitives and alleviate the effect of semantic gaps, and enhances those important ones to better compose novel classes, which consistently achieves superior performance in few-shot image and video recognition. Moreover, the ablation study demonstrates the rationale and positive effect of each module, and reveal insights of our method via visualizing shared primitives between known and novel classes.  

\section{Acknowledgments}
This work is partially supported by grants from the National Key R\&D Program of China under grant 2017YFB1002400, the National Natural Science Foundation of China under contract No. 61825101, No. U1611461, No. 61902131), the Program for Guangdong Introducing Innovative and Enterpreneurial Teams (Grant No. 2017ZT07X183), the Fundamental Research Funds for the Central Universities (Grant No. 2019MS022), China Scholarship Council (CSC) Grant \#201906010176, and grants from NVIDIA NVAIL program and the NVIDIA SaturnV DGX-1 AI supercomputer. 

\bibliographystyle{ACM-Reference-Format}
\bibliography{zoilsen}

\end{document}